# Spiking Neural Network Integrated Circuits: A Review of Trends and Future Directions

*Arindam Basu[1], Charlotte Frenkel[3], Lei Deng[2], Xueyong Zhang[4]

1 City University of Hong Kong, Hong Kong

2 Center for Brain Inspired Computing Research (CBICR), Tsinghua University and Chinese Institute for Brain Research, Beijing, China

3 Institute of Neuroinformatics, UZH and ETH Zurich, Switzerland

4 Nanyang Technological University, Singapore

* Authors listed alphabetically

*Abstract*—In this paper, we reviewed Spiking neural network (SNN) integrated circuit designs and analyzed the trends among mixed-signal cores, fully digital cores and large-scale, multi-core designs. Recently reported SNN integrated circuits are compared under three broad categories: (a) Large-scale multi-core designs that have dedicated NOC for spike routing, (b) digital single-core designs and (c) mixed-signal single-core designs. Finally, we finish the paper with some directions for future progress.

## I. INTRODUCTION

The rapid growth of deep learning, spurred by its successes in various fields ranging from face recognition [1] to game playing [2], has also triggered a growing interest in the design of specialized hardware accelerators to support these algorithms. This specialized hardware targets one of two categories—either operating in datacenters or on mobile devices at the network edge. While energy efficiency is important in both cases, the need is extremely stringent in the latter class of applications due to limited battery life. Several techniques have been used in the past to improve the energy efficiency of these accelerators [3], including reducing off-chip DRAM access, managing data flow across processing elements as well as in-memory computing (IMC) by exploiting analog processing of data within digital memory arrays [4].

A very different approach to improving energy efficiency can be obtained by taking inspiration from the brain and adopting a "neuromorphic" approach. Biological neurons perform local information processing with tightly coupled memory and communicate by passing 1-bit messages or spikes [5], resulting in extremely sparse operation and concomitantly high energy efficiencies. It is estimated that the brain dissipates only about 20W while performing its myriad functions which often surpass the capabilities of today's machine leaning algorithms such as learning from few examples, understanding speech in noisy environments etc. Hence, there has been an increasing amount of interest in exploring a class of brain-inspired neural networks termed spiking neural networks (SNNs). These SNN integrated circuits promise energy efficiency gains primarily by exploiting the event-driven nature of SNN operation where there is no explicit clocked operation unless there are inputs to be processed. In addition, spatio-temporal sparsity of neuronal activation enables further reduction in operations and memory access.

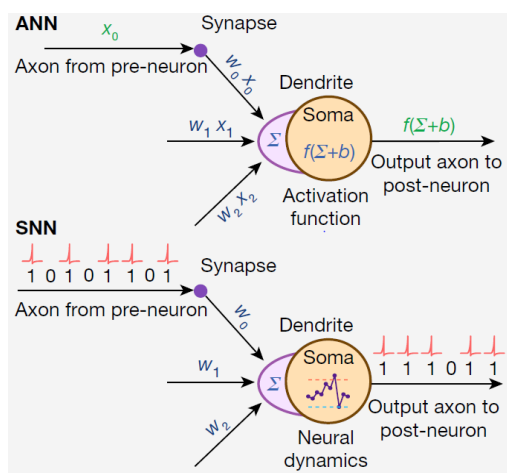

Fig. 1. Spiking neuron vs a traditional artificial neuron where the temporal neuronal dynamics is absent (adapted from [11]).

Fig. 1 depicts the block diagrammatic view of a popular spiking neuron model and compares it with a traditional ReLU neuron. In the simplest form of a spiking neuron, a leaky integrate and fire (LIF) neuron integrates inputs over time in a variable Vmem representing the membrane potential of a biological neuron. This value leaks towards a resting potential at a constant rate depending on a leak conductance or current source. The neuron signals a binary 1-bit output only if the integrated value exceeds a threshold. After such a firing event, Vmem is reset to a fixed value and, depending on the model details, it may be prevented from further integrating inputs during a refractory period. This temporal integration aspect is entirely missing in traditional artificial neurons and hence it is expected that SNNs may be better suited to process temporally varying signals [6,49] such as speech, visual tracking, biomedical sensory data etc.

In this paper, we review the recent progress in developing dedicated hardware to implement SNNs. While other reviews have focused on circuit aspects of traditional ANN accelerators [3,6,42], such a discussion is less common for

Table I. Comparison Table for SNN integrated circuits with task-independent metrics. Columns shaded in light gray have been normalized to a 40-nm CMOS node, and optionally to a 0.9-V supply voltage.

| Design, reference | Vdd (V) | In-memory | Mixed-signal | Process (nm) | # of neurons | # of synapses | Area* (mm²) | Normalized synaptic density (Msyn/mm²) | Energy efficiency (GSOPS/W) (tot / dyn) | Normalized energy efficiency (process & $V_{DD}$) | Normalized energy efficiency (process only) |
|---|---|---|---|---|---|---|---|---|---|---|---|
| Large Scale, multi-core ||||||||||||
| SpiNNaker [8,64] | 1.2 | ✗ | ✗ | 130 | 18k | 18M | 88.4 | 1.912 | 0.033 / 0.088 ⁂ | 0.19 | 0.108 |
| TrueNorth [9,66] | 0.775 | ✗ | ✗ | 28 | 1M | 256M | 413 | 0.304 | 400 / - | 207.6 | 280 |
| Loihi [10] | 0.75 | ✗ | ✗ | 14 | 128k | 128M | 60 | 0.265 | - / 42.4 | - | - |
| Tianjic [11,67] | 0.85 | ✗ | ✗ | 28 | 39k | 9.75M | 14.4 | 0.311 | **649 / -** | **405.22** | **454.3** |
| IFAT [12] | 1.2 | ✗ | ✓ | 90 | 64k | 256k ‡ | 13.3 | 0.024 | 45.4 / - ‡ | 181.82 | 102.27 |
| DYNAPs [13] | 1.8 | ✓ | ✓ | 180 | 1k | 64k | 38.6 | 0.034 | 33.3 / 7460 | 600 | 150 |
| MorphIC [15] | 0.8 | ✗ | ✗ | 65 | 2k | 528k | 2.86 | 0.488 | 19.6 / 33.3 | 25.18 | 31.86 |
| Chen et al. [16] | 0.53 | ✗ | ✗ | 10 | 4k | 1M | 1.72 | 0.036 | 263.2 / - | 22.82 | 65.79 |
| Cho et al. [17] | 0.7 | ✗ | ✗ | 40 | 2k | 149k | 2.56 | 0.058 | 169.5 / - | 102.53 | 169.5 |
| Novena [18] | 0.5 | ✗ | ✗ | 40 | 2k | 256k | 5 | 0.051 | 208.3 / - | 64.3 | 208 |
| Fully-digital, single-core ||||||||||||
| Wang et al. [19] | 0.5 | ✗ | ✗ | 65 | 650 | 65k | 1.99 | 0.086 | 666.7# | 334.36 | 1083.3 |
| μBrain [20] | 1.1 | ✗ | ✗ | 40 | 336 | 36k | 1.42 | 0.025 | 38.5 / 54.9 | 57.45 | 38.46 |
| Seo et al. [21] | 0.53 | ✗ | ✗ | 45 | 256 | 64k | 0.78* | 0.104 | - | - | - |
| ODIN [22] | 0.55 | ✗ | ✗ | 28 | 256 | 64k | 0.086 | 0.365 | 78.7 / 119 | 20.58 | 55.12 |
| Knag et al. [23] | 1 | ✗ | ✗ | 65 | 256 | 128k | 3.06 | 0.11 | - ** | - ** | - ** |
| Kim et al. [24] | 1 | ✗ | ✗ | 65 | 256 | 83k | 1.8 | 0.12 | - *** | - *** | - *** |
| Park et al. [25] | 0.8 | ✗ | ✗ | 65 | 410 | 194k | 10.1 | 0.051 | **<3400$ / -** | **4365.43** | **5525** |
| IMPULSE [26] | 0.85 | ✓ | ✗ | 65 | 192 | 1.5k | 0.089 | 0.045 | 990 / - | 1434.97 | 1608.75 |
| Mixed-signal, single-core ||||||||||||
| Neurogrid [7] | 3 | ✗ | ✓ | 180 | 64k | 256k ‡ | 149 | 0.035 | 1.06 / - □ | 53.13 | 4.78 |
| ROLLS [27] | 1.8 | ✓ | ✓ | 180 | 256 | 128k | 44 | 0.059 | - / 13000 | - | - |
| HICANN [28] | - | ✗ | ✓ | 180 | 512 | 112k | 49 | 0.046 | 10 / - | - | 45 |
| Mayr et al. [29] | 1 | ✗ | ✓ | 28 | 64 | 8k | 0.36 | 0.01 | 1.18 / - | 1.017 | 0.824 |
| Brink et al. [30] | 2.4 | ✓ | ✓ | 350 | 100 | 30k | 21.7 | 0.106 | 100 / - | 6222.2 | 875 |
| Buhler et al. [31] | 0.9 | ✗ | ✓ | 40 | 512 | N/A | 1.3 | - | 3430$ | 3430 | 3430 |
| BrainDrop [32] | 1 | ✗ | ✓ | 28 | 4k | 16M | 0.65 | - | 2630 / - | 2274 | 1842 |
| Yan et al. [33] | 1.8 | ✓ | ✓ | 150 | 256 | 64k | 0.6 | 1.173 | 3890 / - ◊ | 58366 | 14591 |
| Wan et al. [34] | 1.8 | ✓ | ✓ | 130 | 256 | 64k | 1.79 | 0.378 | **74000 / - ◊** | **96200** | **24050** |
| MNIFAT [35] | 5 | ✗ | ✓ | 500 | 4k | 1 ‡ | 9 | 0.069 | 2.78 / - ‡ | 1071.7 | 34.72 |
| HICANN-X [36] | 1.2 | ✗ | ✓ | 65 | 512 | 128k | 27.9 | 0.012 | 1280 / - | 3703.7 | 2083.3 |
| Miscellaneous (FPGA designs, pre-silicon results, building blocks) ||||||||||||
| DeepSouth [37] | - | ✗ | ✗ | - | **20M** | **4T** | - | - | - | - | - |
| Minitaur [38] | - | ✗ | ✗ | - | 64k | 16M | - | - | 0.012 / - | - | - |
| SPOON [39] | 0.6 | ✗ | ✗ | 28 | 8k | 64k | 0.32 | 0.098 | 147 / - | 45.7 | 102.9 |
| Pu et al. [14] | 0.85 | ✗ | ✗ | 40 | 4k | 1M | 4.5 | 0.222 | 222.2 / - | 198.22 | 222.2 |
| Koo et al. [41] | 1.1 | ✓ | ✗ | 28 | 0 | 64k | 0.266 | 0.118 | - | - | - |

* Pads excluded. * Area of the base design. ** Energy efficiency of 47.6 pJ/pix. *** Energy efficiency of 5.7 pJ/pix. # Detail of total or dynamic-only energy dissipation not provided. $ Value in GOPS/W, detail of relation between GOPS/W and GSOPS/W not provided. Forms a higher bound for the number of TSOPS/W.
◊ MAC efficiency taken as a proxy for the SOP efficiency. ⁂ From [74]. ‡ Shared synaptic filters, off-chip weight storage. □ Board-level measurement.

the SNN counterpart. Compared to [68,69], which provide a review about the generic architectural aspects of SNN ICs, we provide more detailed and updated discussion on the circuit aspects. The review of [49], by outlining how the SNN and ANN trends converge, can also come in complement.

The outline of the paper is as follows. Trends of energy efficiency and area efficiency are explored in Section II. Section III presents the key concepts used in the design of digital and mixed-signal neural cores. Section IV discusses various large-scale, multi-core designs with integrated network-on-chip (NoC) for routing spikes. Lastly, Section V presents some discussions about the key future directions in the field.

## II. TRENDS IN SNN ACCELERATORS

Recently reported SNN integrated circuits are compared in Table I under three broad categories: (a) Large-scale multi-core designs that have a dedicated NoC for spike routing, (b) digital single-core designs and (c) mixed-signal single-core designs. Some other designs including FPGA-based ones [37,38,40], pre-silicon results [39,14] or sub-systems of an SNN core [41] are grouped under a miscellaneous category. The data is made available [75] for reference. To facilitate comparison across different designs, process normalization is applied with 40nm as reference. A separate column with both process and power supply voltage (VDD) normalization (0.9V reference) for energy efficiency is also included; however, since the energy efficiency of analog/mixed-signal designs may not scale similarly to digital designs (due to requirements of keeping transistors in saturation, etc.), the energy efficiency is plotted with process normalization only. Only task-independent metrics are compared in this table for the purpose of extracting hardware architectural trends, while task-related benchmarking comparisons are provided in Sections III and IV. Furthermore, throughput-related numbers are not often reported for SNN hardware unlike their ANN accelerator counterparts. This is often due to the input/task-dependent activity of SNNs and hence is difficult to specify. Therefore, we have not included that metric and correspondingly modified the normalized area efficiency metric to be in million synapses per mm$^2$ instead of the more conventional TOPS per mm$^2$.

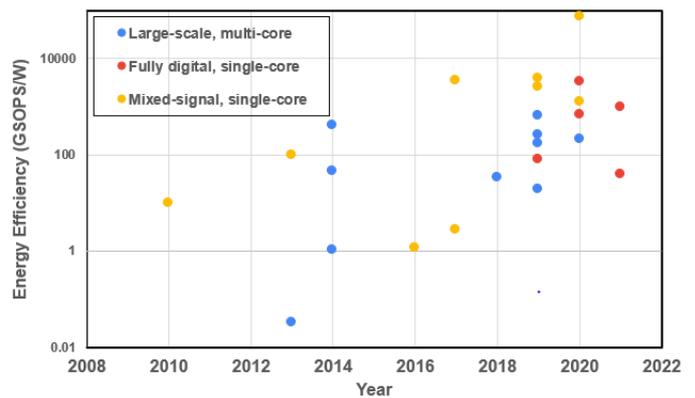

Fig. 2. SNN designs show an increasing energy efficiency trend over time, with an advantage for recent mixed-signal cores.

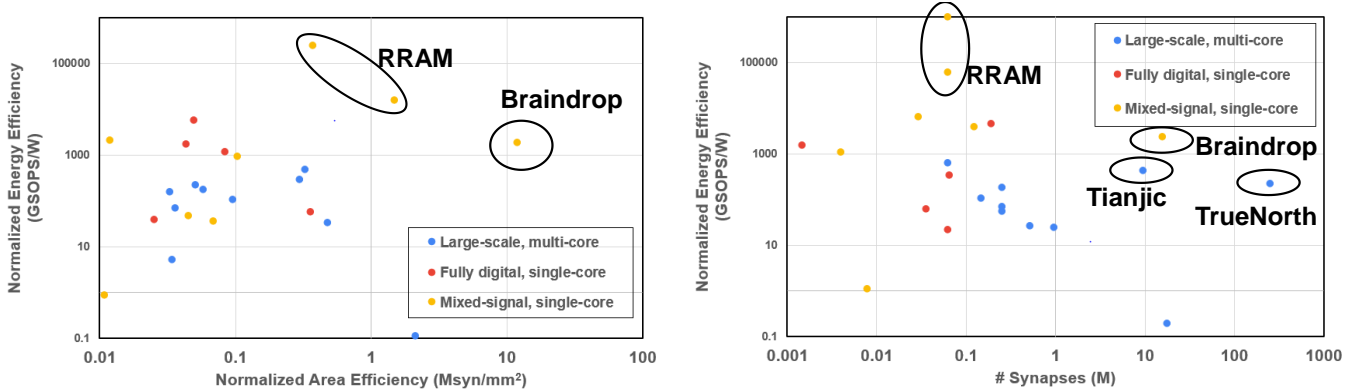

Fig. 3. Normalized energy efficiency plotted against (a) normalized area efficiency and (b) number of million synapses for SNN designs. Mixed-signal designs exhibit an improved energy efficiency with larger synaptic densities.

Moreover, most designs explicitly report energy efficiency in terms of synaptic operations per watt (SOPS/W) which includes communication energy in addition to computation energy (OPS/W) while a few others do not. This makes it difficult to compare designs exactly on the same grounds, but the key trends are not affected much by this discrepancy.

Some key trends emerge as a result of this comparison—Fig. 2 plots the trend of energy efficiency over time. It can be seen that there is a general increase in energy efficiency over time with recent designs approaching 100 TSOPS/W. The highest energy efficiencies are obtained by the mixed-signal cores employing in-memory computation reported with the use of RRAM [33,34] or low-resolution synapses and intrinsic mismatch with sub-threshold analog design [32]. Another interesting trend is observed in the relation between the normalized energy efficiencies and synaptic densities as shown in Fig. 3(a). It is seen that the two metrics are positively correlated exclusively for mixed-signal cores. The probable reason is that both the energy and area costs of DAC and ADCs at the periphery of the memory are amortized better for larger synaptic memory arrays, leading to higher values of both metrics. For digital designs, there is an increased energy cost of accessing a single neural and synaptic information from larger memory arrays with better area efficiency. Also, compared to the ANN counterparts, digital SNN designs need more memory accesses for neuronal state and parameter retrieval as well, exacerbating this effect further. To further explore this hypothesis, we also plot in Fig. 3(b) the normalized energy efficiency against the total number of synapses which is indicative of the total area of the chip.

Indeed, we see that the trend holds true over five orders of magnitude of synaptic numbers and is similar to the energy efficiency vs SRAM size trend of ANN accelerators [6]. Another reason for lower energy efficiency of larger scale designs is the increased energy needed to communicate spike packets over longer distances. This trend suggests using analog cores if the number of synapses per core is higher than ~10k. It also points to the importance of increasing the focus on optimizing the spike communication network infrastructure in future.

Among other designs, [39] presents an online-learning event-driven CNN that exploits single-spike timings for high accuracy as well as energy efficiency in pre-silicon results. Another interesting method of increasing energy efficiency is to reduce synaptic processing time by pipelining and axon skipping for zero inputs. Verified by pre-silicon results [14], this method reduces cycle time and thus has a lower penalty due to static power dissipation. Several designs have also used FPGAs for their implementation. The largest number of neurons and synapses have been implemented in FPGA designs exploiting cortical connectivity styles [37]. Another FPGA design [38] uses the concept of locality in caches to reduce memory fetches from off-chip DRAM.

### III. SINGLE-CORE DESIGNS

Single-core neuromorphic designs implement three main types of network topologies (Table II). (i) Most designs proposed to date implement a crossbar architecture, where an all-to-all neuron connectivity can support fully-connected, recurrent, as well as convolutional layers. However, this support for arbitrary network topologies usually leads to a limited utilization of available synaptic resources, which impedes scalability and leads to a degraded power-performance-area (PPA) tradeoff at the task level. (ii) Locally-competitive-algorithm-based architectures (LCAs) introduce competition between neurons for sparse input encodings, thereby leading to an interesting flexibility-efficiency tradeoff. The first design of [23] was benchmarked for image reconstruction, while the subsequent designs of [24,31] were extended with output classifiers for image recognition. (iii) Recently, multi-layer spiking architectures were introduced and currently offer the best PPA tradeoffs at the task level. However, the flexibility of these designs is reduced as the network topology is not reconfigurable, although μBrain allows generating different network architectures at synthesis time [20]. Despite the inherently time-based event-driven

Table II. Task- and architecture-focused overview of single-core digital and mixed-signal designs.

| Design, publication, reference | Impl., area [◊] (node) | | Topology | On-chip learning (★) | Demonstrated applications | | | |
|---|---|---|---|---|---|---|---|---|
| | | | | | Task (dataset) | Accuracy | E/sample | Throughput [samples/s] |
| *Seo et al.* CICC'11 [21] | Digital 0.78mm² (45nm) | Crossbar | (256)-256 | Stoch. STDP | Pattern recall (custom) Image recog. (custom) | N/A | | |
| ODIN TBioCAS'19 [22] | Digital 0.086mm² (28nm) | | (256)-256 | SDSP | Image recog. (MNIST)* Gestures recog. (EMG [43]) | 91.4% / 84.5% (★) 53.6% | 15nJ 7.4μJ | N/A 42.5 |
| IMPULSE SSCL'21 [26] | Digital (IMC) 0.089mm² (65nm) | | (128)-192 | ✗ | Image recog. (MNIST) Sentiment recog. (IMDB) | (98.96%) [‡] (88.15%) [‡] | N/A | N/A |
| MINIFAT ISCAS'17 [35] | **Mixed** 9mm² (0.5μm) | | (N/A)-4k [∥] | ✗ | Image filtering (custom) | N/A | | |
| *Brink et al.* TBioCAS'13 [30] | **Mixed** 21.7mm² (0.35μm) | | (300)-100 | STDP | N/A | N/A | | |
| *Mayr et al.* TBioCAS'16 [29] | **Mixed** 0.36mm² (28nm) | | (128)-64 | SDSP | N/A | N/A | | |
| ROLLS Frontiers'15 [27] | **Mixed** 44mm² (0.18μm) | | (256)-256 | SDSP | Image recog. (2-class Caltech 101) | N/A | | |
| HICANN ISCAS'10 [28] | **Mixed** 49mm² (0.18μm) | | 2x (224)-256 | STDP | Image recog. (5-class MNIST [44])** | 95% | N/A | N/A |
| HICANN-X arXiv'20 [36] | **Mixed** 27.9mm² (65nm) | | (256)-512 | Flexible | Image recog. (MNIST [45])* | 96.9% | 8.4μJ | 20.8k |
| *Yan et al.* VLSI-C'19 [33] | **Mixed (IMC)** 0.6mm² (0.15μm) | | (256)-256 | ✗ | Image recog. (MNIST) Image recog. (CIFAR10) | 79.1% – (95.5%) [‡] (78.8%) [‡] | 6.0nJ – N/A N/A | 2.5M – N/A N/A |
| *Wan et al.* ISSCC'20 [34] | **Mixed (IMC)** 1.79mm² (0.13μm) | | (256)-256 | ✗ | Image reconstruction (MNIST)*** | 1.91 MSE | N/A | N/A |
| *Knag et al.* JSSC'15 [23] | Digital 3.06mm² (65nm) | LCA | 4x64 | SAILnet | Image reconstruction (custom, 16x16 image patches) | 8.4 x10⁻³ NRMSE [‡] | 12.2nJ [‡] 109nJ (★) [‡‡] | 547k [‡] 62.5k (★) [‡‡] |
| *Kim et al.* VLSI-C'15 [24] | Digital 1.8mm² (65nm) | | 4x64 | SGD (last layer) | Image recog. (MNIST) | 84% – 90% (★) | 27 – 162nJ 94.7μJ (★) | 9.9M – 1.6M 5.5k (★) |
| *Buhler et al.* VLSI-C'17 [31] | **Mixed** 1.3mm² (40nm) | | 8x64 | N/A | Image recog. (MNIST) | 88% (★) | 50nJ | 1.7M |
| *Park et al.* JSSC'20 [25] | Digital 10.1mm² (65nm) | Multi-layer | (784)-200-200-10 | Mod. segr. dendrites | Image recog. (MNIST) | 97.8% (★) | 236nJ [°] 254nJ (★) [°°] | 100k [°] 94.3k (★) [°°] |
| *Wang et al.* ASSCC'20 [19] | Digital 1.99mm² (65nm) | | 256-128-128-128-16 | ✗ | Image recog. (MNIST)* KWS (HeySnips / 4-GSCD)**** | 97.6% 95.8% / 91.8% | 195nJ N/A | 2 N/A |
| μBrain Frontiers'21 [20] | Digital 1.42mm² (40nm) | | rec256-64-16 | ✗ | Image recog. (MNIST)* Gestures recog. (radar, custom) | 91.7% 93.4% | 308nJ 340nJ | N/A |
| Neurogrid PIEEE'14 [7] | **Mixed** 149mm² (0.18μm) | | Flexible (off-chip tree net) (64k neur, 64k syn, off-chip weights) [□] | ✗ | Cortical simulations (custom [52]) | N/A | | |
| Braindrop PIEEE'19 [32] | **Mixed** 0.65mm² (28nm) | | Encode-transform-decode (4k neur, max. 16M syn) | ✗ | Neural Eng. Framework (custom, 2D function fitting / delay line) | 0.05 NRMSE / 0.146 NRMSE | N/A | N/A |

[◊] Pads excluded.  * MNIST dataset shrinked to 16x16 pixels.  ** MNIST dataset shrinked to 10x10 pixels, restricted to digits 0, 1, 4, 6, 7.  *** MNIST dataset shrinked to 15x15 pixels.
**** GSCD dataset restricted to classes "yes", "stop", "right", "off".  [‡] No full on-chip network storage, layers mapped on-chip one at a time.  [‡‡] At 440mV core memory supply.
[‡‡] At 580mV core memory supply.  [°°] At 0.8V, 20MHz.  [∥] Off-chip weight storage; either 2k Mihalas-Niebur or 4k LIF neurons.  [□] Shared synaptic filters.

nature of spike-based processing, all designs in Table II were benchmarked with static stimuli such as images, or with dynamic data pre-processed to static rate-based stimuli. To date, two clear exceptions are Neurogrid [7] and Braindrop [32]. In the former, each chip forms a Neurocore of 256×256 neurons. The Neurogrid is obtained by connecting Neurocores with a flexible tree-based routing scheme at the circuit board level, leading to the demonstration of million-neuron cortical simulations. In the latter, an encode-transform-decode architecture supporting the Neural Engineering Framework (NEF) [46] is implemented, which allows for the deployment of dynamical systems while abstracting out the neuron and synapse implementation details. Overall, within and across the network topologies introduced above, different design strategies are being pursued, from the circuit implementation strategy to the memory architecture.

*Circuit implementation:* From digital to mixed-signal, all circuit implementation strategies are represented in Table II. Most digital designs follow a standard synchronous design flow [21-26], although asynchronous designs with local spike-triggered oscillators were recently introduced [19,20]. This strategy is successfully used in combination with spatio-temporally fine-grained clock and power gating in [19] to achieve jointly high energy efficiency as well low static power needed in always-on applications like keyword spotting. As they best support the event-driven nature of neuromorphic sensors, digital asynchronous spike routing infrastructures are also a standard choice for mixed-signal designs [27,28,30,32,35]. Therefore, the circuit implementation strategies of mixed-signal designs mainly differ at the level of the neuron and synapse building blocks. Subthreshold analog design is the historical approach at the roots of neuromorphic engineering. By operating the MOS transistor in weak inversion, the minority carrier flow in the MOS transistor channel follows a diffusion mechanism, which allows for a direct physics-based emulation of the brain ion channel dynamics in biological time. Key designs in this category include [27,30,32]. On the contrary, above-threshold analog designs operate MOS transistors in strong inversion. As currents are several orders of magnitude higher than in subthreshold designs, above-threshold designs allow reaching acceleration factors of up to 104× compared to biological time. Key examples in this category include the HICANN chips part of the BrainScaleS 1 and 2 wafer-scale systems [28,36]. Finally, an interesting strategy is followed in [29] with a switched-capacitor design. Operating in the charge domain instead of the current domain, the sensitivity to noise, mismatch, and process, voltage and temperature (PVT) variations is reduced compared to subthreshold analog designs. Flexible time constants from biological- to accelerated-time can also be achieved, while the digital control overhead can be reduced to low footprints in advanced technology nodes.

*Memory architecture:* Memory organization is crucial to achieve high energy efficiencies for neural network hardware, both ANNs and SNNs. The highest energy efficiencies [27,33,34] are obtained by designs adopting in-memory computing for the synaptic vector-matrix multiply (VMM) operation. While [33,34] use non-volatile RRAM for synaptic storage, [27] uses capacitors for volatile weight storage with CMOS circuits for bi-stable weight storage. All of them leverage Kirchoff's-current-law-based summation of synaptic outputs. An interesting example of digital in-memory computing is provided in [26] where a 10-T SRAM-based macro fuses neuron Vmem and synaptic weight memories. By reconfiguring peripherals, it can implement accumulate, thresholding, spike-check, and reset operations in-memory, further reducing the need of data movement. Most other digital designs and some analog designs (e.g., [28,29,31,32,36]) use standard SRAM to store synaptic weights which are fetched individually for operation by the neuron circuits. Finally, some designs also rely on off-chip synaptic weight storage (e.g., [7,35]), at the expense of increased energy and latency footprints.

Memory organization also affects on-chip learning and the related trends are also highlighted in Table II. Indeed, since the early developments of neuromorphic silicon devices, synaptic plasticity has been among the key research areas. The first learning rules that were investigated are directly grounded on physiological evidence and focused on spike-based unsupervised learning mechanisms operating locally between two neurons, such as spike-timing-dependent plasticity (STDP) [47] and spike-driven synaptic plasticity (SDSP) [48]. The low computational complexity of STDP and SDSP make them suitable for low-cost implementations in both the analog and the digital domains, although with a limited task performance as these are inherently local mechanisms that do not optimize for a global network error. Therefore, successful task-level on-chip-learning demonstrations follow a more top-down gradient-based approach, while involving different degrees of neuroscience insight [49]. Examples range from non-spike-based vanilla stochastic gradient descent (SGD) in dedicated output classifiers [24] to the SAILnet algorithm, which is a local gradient-based algorithm that minimizes the input representation error, leading to receptive fields similar to those found in the primary visual cortex [23]. Yet, the above-mentioned examples are still working at the level of a single layer. The first demonstration of on-chip bio-inspired multi-layer learning was shown in [25] with a modified implementation of the segregated dendrites algorithm [50]. This algorithm derives from the family of feedback-alignment-based algorithms [51], from which the adaptive spiking convolutional neuromorphic processor SPOON also derives (pre-silicon results in [39]).

## IV. LARGE-SCALE, MULTI-CORE DESIGNS

Multi-core neuromorphic designs can be roughly segmented into two categories. On the one hand, smaller-scale multi-core designs (Table III, bottom) are typically proof-of-concept demonstrations for the scalability of single-core designs. Therefore, these designs usually do not provide well-supported development environments (at the exception of DYNAPs [63]), and their task-level comparison outlines similar conclusions to those drawn for single-core designs in Section III. On the other hand, large-scale multi-core designs (Table III, top) are supported by software and inter-chip interconnect infrastructures scaling to 105 – 109 neurons at the system level. These resources allow for more flexibility, which is demonstrated with applications ranging from larger-scale image recognition datasets to complex tasks spanning autonomous bike driving and cortical simulations. The remainder of this section will thus focus on the key trends for large-scale designs. For the sake of completeness, as a few single-core designs have been demonstrated in large-scale setups (i.e., HICANN in the wafer-scale BrainScaleS [28] and Neurogrid with 1M neurons at the board level [7]), we included them as well in this discussion.

Constructing an aspirational brain-level simulation platform needs powerful and efficient building blocks, especially for large-scale neuromorphic chips. There are several well-known such neuromorphic chips/platforms over the world, e.g., Neurogrid [7], BrainScaleS [28], SpiNNaker [8,64], TrueNorth [9,66], Loihi [10], and Tianjic [11,67]. Neurogrid, TrueNorth, and Loihi lay the foundation for the exploration of brain science in the US, similarly BrainScaleS and SpiNNaker for Europe, and Tianjic for China.

*Design Principles:* Fig. 4 illustrates the typical architecture used by multi-core neuromorphic chips, comprised of many neurosynaptic cores connected by routers. All cores usually work independently, but use handshaking signals to synchronize and share data periodically, which forms a decentralized dataflow. The design of large-scale neuromorphic chips is quite different from the design of smaller ones. First, the scalability is the most important consideration factor towards brain-scale simulation. Due to

Table III. Task- and architecture-focused overview of multi-core digital and mixed-signal designs, with single-chip benchmarking results.

| Design, publication(s), reference | Impl., area◊ (node) | On-chip learning (★) | Demonstrated applications | | | | |
|---|---|---|---|---|---|---|---|
| | | | Task (dataset) ‡ | Mapped topology | Accuracy | E/sample | Throughput [samples/s] |
| SpiNNaker JSSC'13 PIEEE'14 [8,64] | Digital 88.4mm² (130nm) | Flexible | Image recog. (MNIST [53])* Cortical simulations (custom [54])* | 784-500-500-10 N/A | 95% N/A | 6mJ N/A | 50 N/A |
| SpiNN 2 prot. arXiv'21 [65] | Digital 9mm² (22nm) | Flexible | Keyword spotting (custom [55]) ⁺ | 390-256-256-29 ‡ | ~93.8% ‡ | 7.1µJ ‡ | 1k |
| TrueNorth Science'14 TCAD'15 [9,66] | Digital 413mm² (28nm) | ✗ | Image recog. (MNIST/CIFAR10 [56,57]) Phoneme recog. (TIMIT class [57]) Gesture recog. (IBM DVS [58]) | 2- /15-layer CNN 15-layer CNN 16-layer CNN | 92.7% / 83.4% 79.2% 91.8% | 268nJ / 163µJ 57.6µJ >19mJ °° | 1k / 1.25k 2.61k N/A |
| Loihi Micro'18 [10] | Digital 60mm² (14nm) | Flexible | Keyword spotting (custom [55]) ⁺ Gesture recog. (IBM DVS [59]) ** | 390-256-256-29 6-layer CNN | 93.8% 96.2% | 270µJ 2.5mJ | 296 45 |
| Tianjic Nature'19 JSSC'20 [11,67] | Digital 14.4mm² (28nm) | ✗ | Image recog. (MNIST / CIFAR10) Auto. bike driving (image sound and control, custom) | LeNet / VGG-8 5 // multi-layer nets | 99.48% / 93.52% N/A | 76.6µJ / 4.2mJ N/A | 2.1k / 1.75k N/A |
| IFAT BioCAS'14 [12] | **Mixed** 13.3mm² (90nm) | ✗ | N/A **** | N/A | | | |
| DYNAPs TBioCAS'18 [13] | **Mixed** 38.6mm² (0.18µm) | ✗ | Gesture recog. (EMG, custom [62]) *** | 8-192(-3) ‖ | 74% | 100µJ | 2 |
| MorphIC TBioCAS'19 [15] | Digital 2.86mm² (65nm) | Stoch. SDSP | Image recog. (MNIST) Gesture recog. (DVS [43]) | 4x (196-500-10) 4x (400-210-5) | 95.9% 85.1% | 21.8µJ 57.2µJ | 250 58 |
| *Chen et al.* JSSC'19 [16] | Digital 1.72mm² (10nm) | STDP | Image reconstruction (custom) Image recog. (pre-proc MNIST)***** Image recog. (MNIST) | 1024-1024 RBM (★) 236-20 (★) 784-1024-512-10 | 0.036 RMSE (★) 88% (★) 97.9 | N/A 1µJ 1.7µJ | N/A 6.25k N/A |
| *Cho et al.* CICC'19 [17] | Digital 2.56mm² (40nm) | ✗ | Image reconstruction (custom) Image recog. (MNIST) | N/A Rec. layer + classifier | 0.076 NRMSE 91.6% | N/A | N/A |
| Novena ASSCC'20 [18] | Digital 5mm² (40nm) | ✗ | Image recog. (MNIST) | 784-128-128-10 | 97.89% | N/A | N/A |

◊ Pads excluded. ‡ Non-exhaustive list highlighting key demonstrations. * Several MNIST experiments are available for SpiNNaker, the one with detailed single-chip power and latency measurements is provided. The cortical simulations, as for most other SpiNNaker experiments, involves a multi-chip setup. ** Other experiments reported in [60].
*** Image recognition experiments from a 9-chip board are also reported in [13]. **** Benchmarking on an earlier 0.5µm version of IFAT is available in [61].
***** MNIST dataset pre-processed off-chip with Gabor filtering and pooling. ‡ Non-spiking activations. Difference in accuracy results assessed as "negligibly small" compared to the Loihi baseline. Dynamic energy only, static power contributions excluded.
⁺ Pre-processed off-chip to compute MFCC coefficients. °° Only the energy until the first output spike is available, not the energy per full classification. ‖ Off-chip silicon fabrication limits, currently a single chip is impossible to accommodate a full brain, let alone a single region. Scaling a chip up to a board, to a server, and finally to a brain simulator, is an inevitable path. Multi-core architectures with a connectivity infrastructure based on decentralized routers are widely adopted by neuromorphic chips and guarantee the basic scalability; meanwhile, the high communication bandwidth of inter-chip, inter-board, and inter-server interfaces is increasingly becoming the performance bottleneck of a neuromorphic system.

Second, the tradeoff between flexibility and reliability is another consideration factor in the design of such large-scale chips. Interestingly, compared to large-scale designs, in smaller-scale designs we can see more extensive explorations of unconventional neuron models, learning rules, implementation technologies, etc. Sometimes, strong flexibility comes at the expense of reliability. For a large system, the priority of ensuring a predictable operation is higher than seeking for fancy but riskier functionalities. Yet, in order to allow for flexibility in neuroscience-oriented exploration, a suitable balance needs to be found for the flexibility-reliability tradeoff.

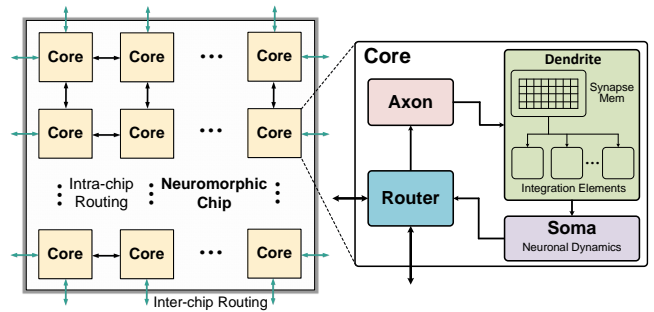

Fig. 4: Generic architecture of multi-core neuromorphic

Third, programmability is becoming a key factor for winning the competition in real-world applications. In early years, the programming of neuromorphic chips relied heavily on human efforts and knowledge of implementation details, which impeded industrialization. To push neuromorphic developments out of the lab environment, the development of efficient programming frameworks and model compilers is attracting an increasing attention, from the early Corelet for TrueNorth [70] to the latest Lava for Loihi 2 [71].

Finally, compatibility is becoming a serious point in constructing large-scale neuromorphic systems, which was usually overlooked in early designs. Currently, the ecology of neuromorphic computing is not mature, thus a bare neuromorphic system without common accelerators and interfaces is difficult to use in practice. At the chip level, adding extra IPs such as image/video codecs is helpful for accelerating real-world applications. At the system level, heterogeneous computing with other devices like CPUs, GPUs, and FPGAs, can be commonly seen. Therefore, adding the support of standard communication interfaces such as Serdes, PCIe, and Ethernet is meaningful even though not directly linked with neuromorphic computing.

Table IV. Functionality comparison of large-scale neuromorphic chips.

|  | Signals | Topology | Modelling | Learning |
|---|---|---|---|---|
| Neurogrid [7] | Mixed | Tree | SNNs | ☒ |
| BrainScaleS [28] | Mixed | N. A. | SNNs | ☑ |
| SpiNNaker [8,64] | Digital | Hexagon | Hybrid | ☑ |
| TrueNorth [9,66] | Digital | 2D Mesh | SNNs | ☒ |
| Loihi [10] | Digital | 2D Mesh | SNNs | ☑ |
| Tianjic [11,67] | Digital | 2D Mesh | Hybrid | ☒ |

Table V. Scale comparison between large-scale neuromorphic chips. All area values are for single chips, excluding pads.

|  | Process | #Neurons | #Synapses | Area (mm$^2$) |
|---|---|---|---|---|
| Neurogrid [7] | 180 nm | 64k | 256k | 149 |
| BrainScaleS [28] | 180 nm | 512 | 112k | 49 |
| SpiNNaker [8,64] | 130 nm | 18k | 18M | 88.4 |
| TrueNorth [9,66] | 28 nm | 1M | 256M | 413 |
| Loihi [10] | 14 nm | 128k | 128M | 60 |
| Tianjic [11,67] | 28 nm | 39k | 9.75M | 14.44 |

Table VI. Performance comparison between large-scale neuromorphic chips. ★: typical value; ↑: high value.

|  | Throughput | Power | Energy Efficiency |
|---|---|---|---|
| Neurogrid [7] | N. A. | 169 mW | 1.1 GSOPS/W |
| BrainScaleS [28] | N. A. | N. A. | 10 GSOPS/W |
| SpiNNaker [8,64] | 64 MSOPS | 1 W | 64 MSOPS/W |
| TrueNorth [9,66] | 3 GSOPS (★) 58 GSOPS (↑) | 65 mW (★) 145 mW (↑) | 46 GSOPS/W (★) 400 GSOPS/W (↑) |
| Loihi [10] | N. A. | N. A. | <42.4 GSOPS/W |
| Tianjic [11,67] | 608 GSOPS | 937 mW (SNN mode) | 649 GSOPS/W |

*Implementation Technology:* We summarize the functionality comparison of the above large-scale neuromorphic chips in Table IV. Neurogrid and BrainScaleS exploit analog signals to model neuronal dynamics while using digital signals to connect neurons and convey event-driven packets. The analog current-voltage characteristics of transistors operated in the sub-threshold regime provide a natural match with the complex biophysics of neuronal and synaptic activations, and analog computing can respond to stimulus rapidly with low power consumption, especially in the above-threshold regime with high acceleration factors. As digital signals can guarantee stable data transmission even over long distances, mixed-analog-digital circuits seem a promising candidate for neuromorphic computing. Nevertheless, analog circuits are sensitive to PVT variations, and thus difficult to program and control. For this reason, most large-scale neuromorphic chips, including SpiNNaker, TrueNorth, Loihi, and Tianjic, adopt a fully digital design flow. An exception may be the second revision of the BrainScaleS mixed-signal architecture, which is currently being integrated at wafer-scale levels [36].

Regarding the routing topology between neurons, Neurogrid adopts the tree structure, SpiNNaker uses the hexagonal structure, and others broadly select the 2D-mesh structure. Notice that TrueNorth adopts a simple point-to-point routing on the 2D mesh, while Loihi and Tianjic further extend it by increasing flexibility via multicast routing. Usually, tree-like structures are simple and do not suffer circle-aware deadlocks, although they are susceptible to system crashes when a routing path breaks. In contrast, grid-like structures are quite scalable and offer high bandwidth and fault tolerance, which is widely adopted in modern neuromorphic designs. Note that, unless pure dimension-ordered routing is followed, the deadlock avoidance issue should be considered in the model compiler if a grid-like routing topology is used [72].

For the neuron model, most neuromorphic chips support only SNNs, while SpiNNaker and Tianjic offer hybrid modelling by additionally supporting ANNs, which generates new opportunities for creating neural models. For the learning ability, BrainScaleS provides a dedicated STDP implementation, Loihi flexibly supports spike-based learning rules, while SpiNNaker offers the highest flexibility with programmable rules. Other designs can only perform inference workloads, which reduces complexity in chip design and system construction. Considering the exploration of synaptic plasticity in large-scale neuromorphic computing models, it is necessary to design neuromorphic chips with a flexible yet efficient learning ability in the future. The next generation of the SpiNNaker and BrainScaleS architectures is currently in development and is aligned with these goals [36,65].

*Performance Comparison:* Table III presents the task-level performance on benchmark datasets for the convenience of readers; however, there is currently no widely accepted benchmark for large-scale neuromorphic chips [73]. Usually, large-scale neuromorphic chips are not designed for specific benchmarks, but for general-purpose neuromorphic computing and exploration. Therefore, we further summarize the general comparison of scale and performance in Table V and Table VI, respectively. Among the aforementioned large-scale neuromorphic chips, TrueNorth integrates the highest number of neurons on a single chip, reaching one million, at the cost of the largest chip area of >4 cm2 in 28-nm CMOS. Owing to the power-efficient sub-threshold analog implementation of Neurogrid and asynchronous circuits of TrueNorth, these two chips demonstrate low power consumption, only about 170 mW per Neurogrid chip and typical 65 mW per TrueNorth chip. Due to the full programmability of the von-Neumann-based SpiNNaker and the low 1-kHz tick frequency of TrueNorth, their synaptic operation (SOP) throughputs are relatively low, only 64 MSOPS per SpiNNaker chip and <60 GSOPS per TrueNorth chip. As Tianjic bridges both artificial and spiking neural models, it can achieve much higher operation throughputs with >600 GSOPS, without the need to strictly follow the slow biological time constants, which comes at the expense of a higher >900 mW power consumption, without power-efficient asynchronous circuits. Considering both the operation throughput and power consumption, TrueNorth and Tianjic demonstrate high energy efficiencies with 400 GSOPS/W and 650 GSOPS/W, respectively. Neurogrid is limited to 1.1 GSOPS/W due to off-chip routing, and the full flexibility of SpiNNaker limits it to 64 MSOPS/W. Loihi has a best-case efficiency of 42 GSOPS/W, which however excludes the cost of neuron updates as well as static power contributions.

## V. Discussion and Conclusion

In this paper, we reviewed SNN integrated circuit designs and analyzed the trends among mixed-signal cores, fully digital cores and large-scale, multi-core designs. The highest energy efficiency in single-core designs is obtained by in-memory computing approaches using volatile or non-volatile memories for synaptic matrix-vector-multiply operations. This is a promising approach for future designs to explore. Large-scale designs show a trend of reducing energy efficiency with larger number of synapses due to the power overhead of large SRAM memories as well as communication energy to send spike packets over longer distances. While digital single-core designs are lower in energy efficiency than their mixed-signal counterparts for core sizes larger than ~10k synapses, they provide the benefit of automated design and ease of porting across technologies. Combined with local spike triggered oscillators [19,20], this is a good direction for future work to combine the energy efficiency of event-driven systems with the convenience of synchronous digital design. Lastly, SNNs are supposed to be energy-efficient due to their event-driven nature—however, most designs optimize energy per event by operating the neuro-cores at a high speed to amortize the static power dissipation. There is a need to focus on simultaneously low-power and low-energy designs.

In the future, it is essential for the whole community to work on standardizing several aspects for better comparison and benchmarking of designs. First, it is extremely important to develop good benchmarks for SNN designs [73]. Most designs reviewed here have been evaluated on MNIST or CIFAR-10 benchmarks that comprise stationary images converted to spikes artificially. However, SNNs are better suited to process temporally encoded data such as speech or biomedical signals [49]. Hence, some datasets like phoneme classification [57], keyword spotting [55] and gesture recognition [59] are better suited for evaluation of SNN ICs in future. Development of more such datasets is an important area for future research. In relation to this, current SNN ICs do not have a good way to compare throughput unlike their ANN accelerator counterparts. This is largely due to their data-dependent throughput. It is important to specify throughput on standard benchmarks as a means to compare different designs. Lastly, the event communication methodology of address-event representation (AER) is modified slightly for different designs, making it difficult for inter-operation of event-driven sensors and processors from different groups. Developing IEEE standards for harmonizing the packet format and physical channel (number of wires, polarities etc.) would be extremely beneficial for inter-operability of designs.